\documentclass{article}
\usepackage{spconf,amsmath,graphicx}
\usepackage{appendix} 
\usepackage{balance,cite,url,bm} 
\usepackage[ruled, linesnumbered]{algorithm2e}
\usepackage{arydshln}
\usepackage{multirow}
\usepackage{graphicx}
\usepackage{booktabs}

\title{Perception-Distortion Trade-off with Restricted Boltzmann Machines}
\name{Chris Cannella$^{\star}$ \qquad Jie Ding$^{ \dagger}$ \qquad Mohammadreza Soltani$^{\star}$ \qquad Yi Zhou$^{\ddagger}$ \qquad Vahid Tarokh$^{\star}$\thanks{This work was supported in part by the Office of Naval Research Grant No. N00014-18-1-2244 and DARPA Grant No. HR00111890040.}}

			\address{$^{\star}$ Department of Electrical and Computer Engineering, Duke University \\
			    $^{\dagger}$School of Statistics, University of Minnesota \\
			    $^{\ddagger}$ Department of Electrical and Computer Engineering, University of Utah}
\begin{document}
\nobalance
%
\maketitle
\begin{abstract}
In this work, we introduce a new procedure for applying Restricted Boltzmann Machines (RBMs) to missing data inference tasks, based on linearization of the effective energy function governing the distribution of observations.  We compare the performance of our proposed procedure with those obtained using existing reconstruction procedures trained on incomplete data. We place these performance comparisons within the context of the perception-distortion trade-off observed in other data reconstruction tasks, which has, until now, remained unexplored in tasks relying on incomplete training data.  
\end{abstract}
\section{Introduction}
\label{sec:intro}
\vspace{-0.05in}
As data complexity keeps increasing in modern machine leaning tasks, the likelihood of encountering a  failure to collect authentic values of interest also increases. For example, we might expect to encounter sensor malfunctions in a wireless sensor network at a rate proportional to the size of the network. Therefore, there is a growing need to develop machine learning techniques that enable satisfactory training and inference from incomplete data.  Imputation, where missing data values are filled with suitable values inferred from observations, represents a promising technique for extending machine learning methods to handle missing data. 

Given their explicit representation of underlying data distributions, Restricted Boltzmann Machines (RBMs) are an appealing choice for imputing missing values.  With a well trained RBM, the conditional probabilities of the missing values given the observed values remain accessible via either direct calculation (in a theoretical sense) or indirect Gibbs sampling.  A variety of training and imputing procedures have been proposed to allow the application of RBMs to handle  missing data, with various computational costs.  In this work, we propose a new technique for applying RBMs to missing data tasks, which significantly improves imputation performance over the existing comparable approach.

One of the key challenges in employing imputation for decision and classification tasks is ensuring that the information accessible within the original data is not significantly altered after filling unobserved entries with definitive numerical values.  For critical tasks, it would be hazardous to rely on a method of imputing optimized to produce convincing, rather than correct, inferrences. A perception-distortion trade-off has been demonstrated and explored in contexts where complete data is available during training, yet its impact on tasks where training data is also incomplete has not been evaluated.  We therefore also use this introduction of the Linearized Marginal RBM to investigate the perception-distortion trade-off in the context of missing data tasks.

\section{Related Work}
\vspace{-0.05in}
Deep learning approaches have achieved excellent performance in image inpainting tasks.  Techniques like Pathak et al's context encoders \cite{pathak2016context}, Yang et al.'s multi-scale patch synthesis \cite{yang2017high}, Iizuka et al.'s global and local consistency discriminators \cite{iizuka2017globally}, and Yu et al.'s contextual attention \cite{yu2018generative}, are adept at completing images with missing patches.  However, these methods require complete training data with known ground truth and do not address how we should proceed in tasks where training data is also incomplete.

Salakhutdinov et al. \cite{salakhutdinov2007restricted}, introduced a  straightforward methodology for applying RBMs to missing data tasks, wherein parameters of the RBM associated with missing inputs are zeroed during training and imputation.  This method is very similar to the Dropout RBM \cite{srivastava2014dropout}, with visible units dropped rather than hidden units, and enjoys the advantage of imposing no constraints on the type of RBM being trained and only light constraints on the  parameter update method. More computationally involved procedures for handling missing data imputation with RBMs have also been introduced.  Salakhutdinov et al. \cite{salakhutdinov2007restricted} further expanded their original method by adding data missingness indicators as additional visible units.  Zeiler et al. \cite{zeiler2009modeling} investigated Gibbs sampling and energy descent approaches to missing data imputation for Gaussian-Bernoulli RBMs.  Liang et al. \cite{wu2019accelerate} propose a method for training and employing RBMs on missing data using psuedo-completion of missing values and inference from persistent hidden layer states to perform imputation.  Approaches for training conditional RBMs, such as Ping and Ihler's belief propagation technique \cite{ping2017belief}, may be adapted to data imputation tasks, though with high computational cost for both training and inference.

Recently, Generative Adversarial Network (GAN) based approaches for imputation tasks have been introduced and have shown good performance on image inpainting problems. GANs rely on the implicit sampling of the probability distribution of input data through a \emph{zero-sum} game between a generator-discriminator pair.  Boras et al.'s AmbientGAN \cite{bora2018ambientgan} provides a method of training generative models from data affected by known corruption mechanisms (including missing data), though it does not aim to perform the reconstruction of corrupted measurements. Soltani et al. \cite{DemixingGAN2019Asilomar, DemixingGAN2019ICLRWS} used GANs to learn from and reconstruct additively corrupted images.  Yoon et al.'s GAIN \cite{yoon2018gain} trains a single generator-discriminator pair from incomplete data intended to model the data generating probability distribution. Li et al's MisGAN \cite{li2019misgan} trains a second generator-discriminator pair for modeling the missingness process.  

Finally, the performance of image imputation and reconstruction approaches has been studied in a general perception-distortion trade-off framework. In particular,  Blau and Michaeli \cite{blau2018perception} derived a perception-distortion trade-off between the perceptual quality and inferrence accuracy for data completion tasks, explaining prior observations that optimizing for one measure leads to decreased performance with respect to the other.  The investigation of this trade-off has centered on complete data contexts, though it must be present and may be more pronounced in settings where reconstruction mechanisms must be constructed based solely on examples with missing data.  The properties of this trade-off in missing data contexts certainly merit exploration, given their potential impact on our understanding of the achievable performance of inference methodologies.

\section{The Linearized Marginal RBM (LM-RBM)}
\vspace{-0.05 in}
In this work, we consider the task of image imputation using binary-binary RBMs, with the joint density function
$
P(\mathbf{v}, \mathbf{h};\theta) = e^{-E(\mathbf{v}, \mathbf{h};\theta)} / Z
$
that involves a normalizing constant $Z$ and an energy function: 
\begin{equation*}
E(\mathbf{v}, \mathbf{h};\theta) = -\mathbf{b}^{T}\mathbf{v} - \mathbf{v}^{T}W\mathbf{h} - \mathbf{c}^{T}\mathbf{h}.
\end{equation*}
Here, $\bm v \in \{0,1\}^n$ denotes $n$ visible units and $\bm h \in \{0,1\}^m$ denotes $m$ hidden units.
Let $\mathbf{v}_{o}$ and $\mathbf{v}_{m}$ denote the observed and missing visible units. 
The marginal probability density function involving observed visible units and hidden units is given by:
\begin{equation*}
    P(\mathbf{v}_{o}, \mathbf{h};\theta) 
    = \sum_{\mathbf{v}_{m}}\frac{e^{-E(\mathbf{v}, \mathbf{h};\theta)}}{Z} .
\end{equation*}
In general, the marginal probability $P(\mathbf{v}_{o}, \mathbf{h};\theta)$ is not of the same functional form as the complete density function $P(\mathbf{v}, \mathbf{h};\theta)$.
However, it admits an observed energy function in the form of
$
     P(\mathbf{v}_{o}, \mathbf{h};\theta) 
   =  e^{-E_{o}(\mathbf{v}_{o}, \mathbf{h};\theta)} / Z_{o}
$,
where $Z_{o}$ is a normalizing constant. For binary-binary RBMs,
\begin{algorithm}[tb]
\SetInd{0.0cm}{0.2cm}
\KwIn{RBM parameters $\mathbf{b}$, $\mathbf{c}$, and $W$;
List of data and mask pairs $(\mathbf{v}_{i}, \mathbf{m}_{i})$, $batch$; Learning rate $\eta$; Observation probabilities $\mathbf{\lambda}$}
$\Delta \mathbf{b} \leftarrow \mathbf{0}$; $\Delta \mathbf{c} \leftarrow \mathbf{0}$; $\Delta W \leftarrow 0$;\\
 \For{$(\mathbf{v}, \mathbf{m})$ $\mathbf{in}$ $batch$}{
  $ \  \ \mathbf{b}_{m} \leftarrow \sigma(\mathbf{b})\odot\mathbf{m}$; (where $\odot$ denotes the Hadamard product)
  $\mathbf{h} \sim \sigma((\mathbf{v}\odot(1-\mathbf{m}) + \mathbf{b}_{m})W + \mathbf{c}))$\;
  $\mathbf{v}' \sim \sigma(W\mathbf{h} + \mathbf{b})$\;
  $\mathbf{h}' \sim \sigma((\mathbf{v}'\odot(1-\mathbf{m}) + \mathbf{b}_{m})W + \mathbf{c})$\;
  $\Delta \mathbf{c} \mathrel{{+}{=}} \eta(\mathbf{h} - \mathbf{h}')$;
  $\Delta \mathbf{b} \mathrel{{+}{=}} \eta(\mathbf{v} - \mathbf{v}')\odot(1-\mathbf{m})$\;
  $\Delta W \mathrel{{+}{=}} \eta((\mathbf{v} \odot (1-\mathbf{m}))^{T}\mathbf{h} - (\mathbf{v'} \odot (1-\mathbf{m}))^{T}\mathbf{h'})$\;
    $\Delta \mathbf{b} \mathrel{{+}{=}} \eta\mathbf{\lambda} \odot \mathbf{b}_{m} \odot (1 - \sigma(\mathbf{b}))\odot(W(\mathbf{h} -\mathbf{h}'))$\;
  $\Delta W \mathrel{{+}{=}} \eta (\mathbf{\lambda} \odot \mathbf{b}_{m})^{T}(\mathbf{h} -\mathbf{h}')$\;
 }
 $\mathbf{b} \leftarrow \mathbf{b} + \Delta \mathbf{b}$; $\mathbf{c} \leftarrow \mathbf{c} + \Delta \mathbf{c}$; $W \leftarrow W + \Delta W$;
 \caption{LM-RBM Minibatch CD-1 Update}
 \label{algo1}
\end{algorithm}
 \vspace{-0.2cm}
\begin{equation}
E_{o}(\mathbf{v}_{o}, \mathbf{h};\theta) = -\mathbf{b}_{o}^{T}\mathbf{v}_{o} - \mathbf{v}_{o}^{T}W_{o}\mathbf{h} - \mathbf{c}^{T}\mathbf{h} - \mathbf{1}^{T} \zeta(\mathbf{b}_{m} + W_{m}\mathbf{h}),\nonumber
\end{equation}
with $\zeta$ denoting the element-wise softplus function \cite{dugas2001incorporating}.
We propose using a Taylor expansion  to approximate the observed energy function so that the functional form is similar to that of the original energy function.  Linearizing about $W_{m}\mathbf{h}=\mathbf{0}$ (element-wise) yields the following approximated observed energy function, with $\sigma$ denoting the element-wise sigmoid function:
\begin{equation}
\hat{E}_{o}(\mathbf{v}_{o}, \mathbf{h};\theta) = -\mathbf{b}_{o}^{T}\mathbf{v}_{o} - \mathbf{v}_{o}^{T}W_{o}\mathbf{h} - (\mathbf{c} + W_{m}^{T}\sigma(\mathbf{b}_{m}))^{T}\mathbf{h}\nonumber.
\end{equation}
The main difference between this observed energy function and the one used in \cite{salakhutdinov2007restricted} is the extra term $\sigma(\mathbf{b}_{m})^{T}W_{m}\mathbf{h}$.  Our approximated $\hat{E}_{o}$ is then used as the shared-parameter energy function for the corresponding observed RBM. Differing from the method of \cite{salakhutdinov2007restricted}, our $\hat{E}_{o}$ depends on all parameters of the full RBM being trained, allowing parameters associated with missing visible units to be updated to increase the likelihood associated with the observed visible unit states. Experimental results show that this indeed leads to a significant performance gain.  Updating based on $\hat{E}_{o}$ alone can lead to the parameter updates of often missing visible units becoming dominated by contributions from training samples in which their value is unknown.  To avoid this, we scale the update contributions to missing visible unit parameters by the overall probability that those units are observed within the training data.  This heuristic scaling selectively reduces the learning rate for visible units with fewer observations and balances their associated parameters' roles in accurately modeling individual unit behaviors and in modeling the behavior of the visible layer as a whole.  Algorithm~\ref{algo1} outlines how our method may be used with contrastive divergence (CD) \cite{carreira2005contrastive}  to train an RBM on missing data.  The method of \cite{salakhutdinov2007restricted} is obtained by omitting lines 8 and 9 from Algorithm~\ref{algo1}.  Imputation is performed by conditionally sampling hidden layer activations under $\hat{E}_{o}$ and then conditionally sampling a reconstruction of the visible layer under $E$ (as in \cite{salakhutdinov2007restricted}, Eqs. 9 and 10). 

\begin{figure*}[htb]

  \centering
  \centerline{\includegraphics[width=0.97\textwidth]{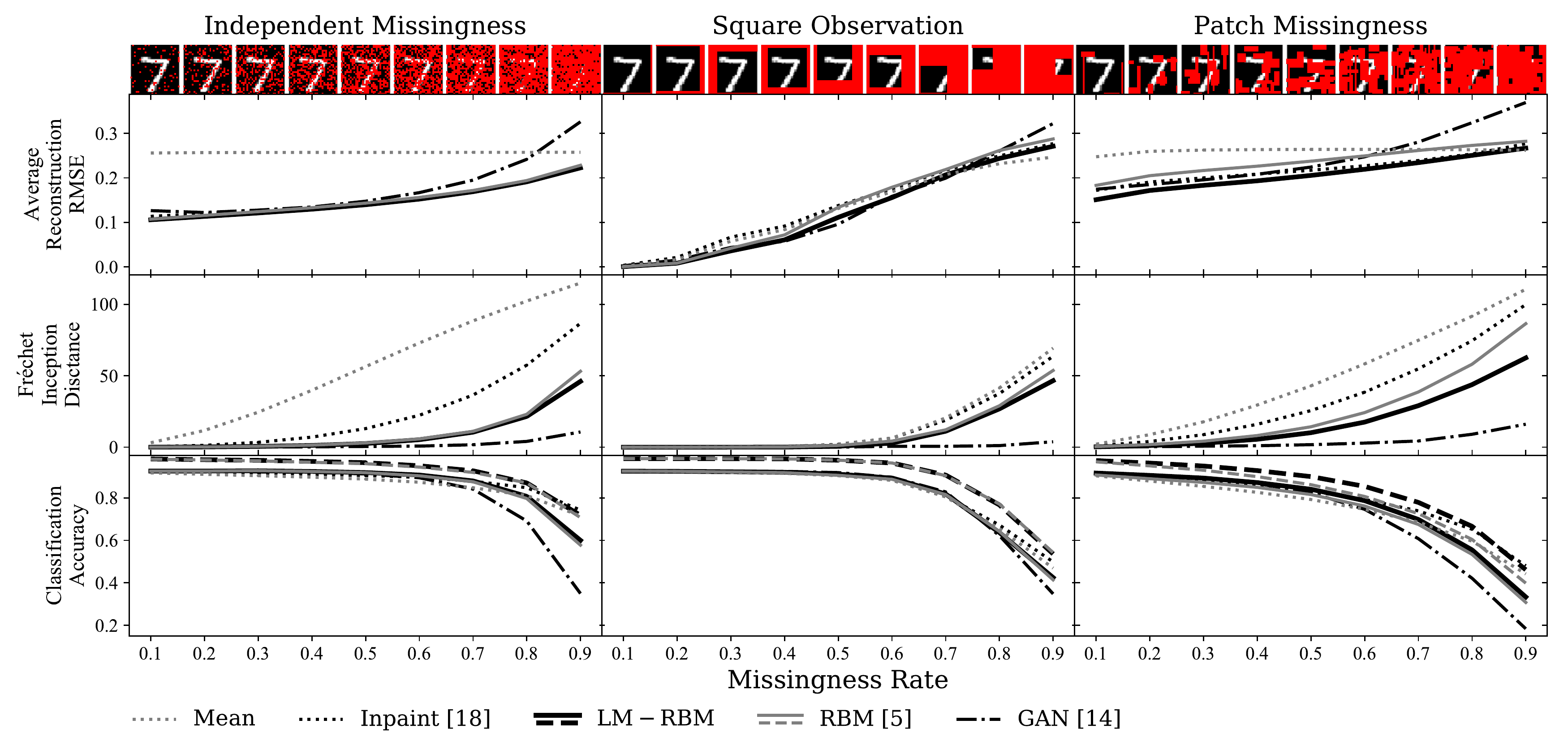}}

\caption{Performance comparison of imputation methodologies completing MNIST digits trained without access to ground truth.}
\label{fig:res}
\end{figure*}
Additionally,~\cite{salakhutdinov2007restricted} proposes to restrict the form of $W$ to the low-rank factorization $W=AB$ (\cite{salakhutdinov2007restricted}, Eq. 14).  We find that this factorization is beneficial when working with missing data and suggest (and adopt) the rule-of-thumb that the rank of this factorization should be fixed to the expected number of observed visible units within the training data.  We justify this heuristic by noting that this is the dimension of the largest linear subspace that we would expect to be fully recoverable through the typical missingness rates in the training data.

\section{Experimental Results}
\vspace{-0.05in}
\setlength{\tabcolsep}{3pt}
\begin{table*}[]
\resizebox{\textwidth}{!}{%
\begin{tabular}{@{}cc@{\hskip 0.2cm}rrrrrrrrrrrrrrrrr@{}}
\toprule
\multicolumn{2}{c@{\hskip 0.2cm}}{} & \multicolumn{5}{c}{Reconstruction RMSE} & \multicolumn{5}{c}{FID} & \multicolumn{7}{c}{Classification Accuracy} \\ \cmidrule(l){3-7} \cmidrule(l){8-12} \cmidrule(l){13-19}
\multicolumn{2}{r@{\hskip 0.2cm}}{Rate} & \multicolumn{1}{c}{Mean} & \multicolumn{1}{l}{\begin{tabular}[c]{@{}c@{}}Inpaint \\ \cite{richard2001fast}\end{tabular}} & \multicolumn{1}{c}{\begin{tabular}[c]{@{}c@{}}GAN \\ \cite{li2019misgan}\end{tabular}} & \multicolumn{1}{c}{\begin{tabular}[c]{@{}c@{}}RBM\\ \cite{salakhutdinov2007restricted}\end{tabular}} & \multicolumn{1}{c}{\begin{tabular}[c]{@{}c@{}}RBM \\ LM\end{tabular}} & \multicolumn{1}{c}{Mean} & \multicolumn{1}{l}{\begin{tabular}[c]{@{}c@{}}Inpaint \\ \cite{richard2001fast}\end{tabular}} & \multicolumn{1}{c}{\begin{tabular}[c]{@{}c@{}}GAN \\ \cite{li2019misgan}\end{tabular}} & \multicolumn{1}{c}{\begin{tabular}[c]{@{}c@{}}RBM\\ \cite{salakhutdinov2007restricted}\end{tabular}} & \multicolumn{1}{c}{\begin{tabular}[c]{@{}c@{}}RBM \\ LM \end{tabular}} & \multicolumn{1}{c}{Mean} & \multicolumn{1}{l}{\begin{tabular}[c]{@{}c@{}}Inpaint \\ \cite{richard2001fast}\end{tabular}} & \multicolumn{1}{c}{\begin{tabular}[c]{@{}c@{}}GAN \\ \cite{li2019misgan}\end{tabular}} & \multicolumn{1}{c}{\begin{tabular}[c]{@{}c@{}}RBM \\ \cite{salakhutdinov2007restricted}\end{tabular}} & \multicolumn{1}{c:}{\begin{tabular}[c]{@{}c@{}}RBM \\ LM\end{tabular}} & \multicolumn{1}{c}{\begin{tabular}[c]{@{}c@{}}RBM $\dagger$ \\ \cite{salakhutdinov2007restricted}\end{tabular}} & \multicolumn{1}{c}{\begin{tabular}[c]{@{}c@{}}RBM $\dagger$ \\ LM\end{tabular}} \\ \addlinespace
\multirow{3}{*}{\rotatebox[origin=c]{90}{I.M.}} & \multicolumn{1}{c@{\hskip 0.2cm}}{0.3} & 0.257(0) & 0.126(0) & 0.128(0) & 0.124(0) & \multicolumn{1}{r}{\textbf{0.121(0)}} & 25(0) & 3.4(0) & \textbf{0.2(0)} & 0.6(0) & \multicolumn{1}{r}{0.5(0)} & 90.5(0) & 91.9(0) & 92.2(1) & 92.8(2) & \multicolumn{1}{r:}{92.8(1)} & 97.4(1) & \textbf{97.7(0)} \\
 & \multicolumn{1}{c@{\hskip 0.2cm}}{0.6} & 0.257(0) & 0.155(0) & 0.167(2) & 0.156(0) & \multicolumn{1}{r}{\textbf{0.152(0)}} & 72(0) & 22(0) & \textbf{0.8(1)} & 5.8(1) & \multicolumn{1}{r}{5.3(0)} & 87.4(1) & 89.7(0) & 89.5(3) & 90.4(1) & \multicolumn{1}{r:}{90.8(1)} & 94.5(1) & \textbf{95.1(1)} \\
 & \multicolumn{1}{c@{\hskip 0.2cm}}{0.9} & 0.257(0) & 0.224(0) & 0.326(4) & 0.228(0) & \multicolumn{1}{r}{\textbf{0.222(0)}} & 115(0) & 87(0) & \textbf{11(1)} & 53(1) & \multicolumn{1}{r}{46(0)} & 71.5(2) & \textbf{74.2(2)} & 35(2) & 57.8(4) & \multicolumn{1}{r:}{60.0(2)} & 70.7(3) & 71.8(4) \\ \addlinespace
\multirow{3}{*}{\rotatebox[origin=c]{90}{S.O.}} & \multicolumn{1}{c@{\hskip 0.2cm}}{0.3} & 0.058(0) & 0.067(1) & 0.044(1) & 0.042(1) & \multicolumn{1}{r}{\textbf{0.036(1)}} & 0.1(0) & 0.1(0) & \textbf{0.0(0)} & 0.0(0) & \multicolumn{1}{r}{0.0(0)} & 92.0(1) & 92.0(1) & 92.5(1) & 92.2(1) & \multicolumn{1}{r:}{92.3(1)} & \textbf{98.4(1)} & 98.4(1) \\
 & \multicolumn{1}{c@{\hskip 0.2cm}}{0.6} & 0.169(0) & 0.175(0) & 0.159(2) & 0.179(1) & \multicolumn{1}{r}{\textbf{0.156(2)}} & 6.3(0) & 6.4(0) & \textbf{1(0)} & 4.3(0) & \multicolumn{1}{r}{2.9(1)} & 88.3(3) & 88.3(2) & 89.7(1) & 88.8(1) & \multicolumn{1}{r:}{89.3(2)} & \textbf{96.4(2)} & 96.4(2) \\
 & \multicolumn{1}{c@{\hskip 0.2cm}}{0.9} & \textbf{0.247(0)} & 0.277(0) & 0.322(1) & 0.287(1) & \multicolumn{1}{r}{0.271(1)} & 69(0) & 64(0) & \textbf{4(1)} & 54(0) & \multicolumn{1}{r}{47(1)} & 47(2) & 49.9(5) & 34.8(5) & 41.4(4) & \multicolumn{1}{r:}{42.5(4)} & \textbf{54.2(5)} & 53.6(4) \\ \addlinespace
\multirow{3}{*}{\rotatebox[origin=c]{90}{P.M.}} & \multicolumn{1}{c@{\hskip 0.2cm}}{0.3} & 0.262(0) & 0.200(0) & 0.196(1) & 0.217(1) & \multicolumn{1}{r}{\textbf{0.183(1)}} & 18(0) & 8.8(0) & \textbf{0.7(1)} & 4.2(1) & \multicolumn{1}{r}{2.8(0)} & 85.4(2) & 87.9(3) & 89.1(2) & 87.3(1) & \multicolumn{1}{r:}{89.3(2)} & 93.0(3) & \textbf{95.0(2)} \\
 & \multicolumn{1}{c@{\hskip 0.2cm}}{0.6} & 0.264(0) & 0.227(0) & 0.247(1) & 0.249(1) & \multicolumn{1}{r}{\textbf{0.219(1)}} & 58(0) & 39(0) & \textbf{2.9(2)} & 24(0) & \multicolumn{1}{r}{18(1)} & 74.6(2) & 79.3(4) & 74.6(4) & 76.2(4) & \multicolumn{1}{r:}{78.7(4)} & 80.7(4) & \textbf{85.4(4)} \\
 & \multicolumn{1}{c@{\hskip 0.2cm}}{0.9} & \textbf{0.262(0)} & 0.277(0) & 0.369(1) & 0.282(0) & \multicolumn{1}{r}{0.267(0)} & 111(0) & 100(0) & \textbf{16(2)} & 87(0) & \multicolumn{1}{r}{63(1)} & 44.2(7) & \textbf{48.0(5)} & 18.4(4) & 30.8(5) & \multicolumn{1}{r:}{33.4(4)} & 39.9(4) & 46.2(4) \\ \bottomrule
\end{tabular}%
}
\caption{Experimentally determined imputation accuracy and perceptual quality measures at representative masking rates.  Value means are reported to at most the first significant digit of standard error. RBM $\dagger$ denotes results for RBM hidden layer.}
\label{tab:my-table}
\end{table*}

In our experimental study, we applied the training and imputation methodologies to the MNIST hand written digit dataset, \cite{lecun1998gradient}, subject to varying missing rates and mechanisms.  We generated training sets of 50,000 digits with missing values and generated similar test sets of 10,000 digits.  For each missingness mechanism and rate, we trained (where applicable) all considered models three separate times on independently generated training sets.  Each trained model was then tested ten separate times on independently generated test sets.  Our reported results are therefore based on thirty distinct pairs of training and testing sets (to record standard errors).

For each missingness mechanism, a range of nine missingness rates were tested between 10\% and 90\%.  Independent missingness (I.M.), where the missingness of individual pixels are i.i.d Bernoulli random variables, and square observation (S.O.), where all pixels are masked except for a randomly placed square of fixed size, were implemented as in \cite{li2019misgan}.  Additionally, patch missingness (P.M.), where a fixed number of randomly placed rectangular regions of pixels are missing, was implemented. We consider two ``untrained'' imputation methodologies.  The first is to replace each missing pixel with its observed mean across the training set.  The second is to iteratively inpaint missing values following the procedure introduced in \cite{richard2001fast}, applying the second diffusion kernel for exactly 200 iterations. For RBM based methodologies, we trained RBMs with 4096 hidden units following both our proposed method and the method from \cite{salakhutdinov2007restricted}.  Both were trained for exactly 500 epochs using mini-batch CD-1 updates with batch sizes of 128 and learning rates of $\eta = 10^{-4}$.  Neither momentum nor explicit regularization was used during the training.

We selected MisGAN as a representative of GAN-based imputation procedures, as it has obtained excellent perceptual quality performance in completing MNIST digits. We trained the convolutional generator and critic variant of MisGAN using the implementation available at \cite{misganrepo}.  We trained all MisGAN instances for exactly 500 epochs with a batch size of 64 using the default training parameters\footnote{$\tau = 0, \alpha=0.1, \beta=0.1, \gamma=0$, $\mathtt{maskgen}=\mathtt{fusion}$, $\mathtt{gp\_lambda}=10$, $\mathtt{n\_critic}=5$, and $\mathtt{n\_latent}=128$, with a three layer fully connected imputer network with 784 units in each layer}.

For evaluation of imputation performance, average image per-pixel reconstruction root mean square error, (RMSE) and the classification accuracy of a multinomial logistic classifier (fit to reconstructions of the training set) on the test set were recorded.  For RBMs, the classification accuracy of a multinomial logistic classifier as applied to inferred hidden layer activations was also recorded.  Fr\'echet inception distance (FID), introduced in \cite{heusel2017gans}, is considered representative of the perceptual quality of the imputed results.  We therefore also record the FID as calculated using the implementation provided within \cite{misganrepo}.  Our experimental results are plotted  in Figure~\ref{fig:res}, with selected numerical results listed in Table~\ref{tab:my-table}.

\section{Conclusion and Future Work}
\vspace{-0.05 in}

We find that, at least with MNIST digits subject to these missingness rates and mechanisms, the relation between perceptual quality and reconstruction accuracy does not differ substantially between the complete and incomplete data settings.  Our results demonstrate the emergence of the perception-distortion trade-off as data missingness rates increase in incomplete data tasks. As less data is made available for training and inference, the correct reconstruction becomes less certain, leading methods optimizing for perceptual quality to produce increasingly distorted reconstructions.  Our results with MisGAN display a prominent degradation of reconstruction accuracy beginning around a 50\% missingness rate for all missingness mechanisms.  

We have limited our comparison with RBM-based methods to the simplest method of \cite{salakhutdinov2007restricted} to isolate the effect of our approximations.  Between the LM-RBM and the method of \cite{salakhutdinov2007restricted}, the LM-RBM offers statistically significant improvements to both reconstruction RMSE and FID across our tests.  LM-RBM also appears to offer improved classification accuracy using its reconstructions, though the classification accuracy results feature greater uncertainty at high missingness rates.  We conclude that, in the terminology of \cite{blau2018perception}, LM-RBM dominates the method of \cite{salakhutdinov2007restricted} on the perception-distortion trade-off at least with respect to the reconstruction RMSE and FID performance measures.  We see this as evidence that the Linearized Marginal approximation improves imputation performance in these settings.

As the LM-RBM consistently offers reconstructions with the lowest RMSE and MisGAN offers those with the lowest FID, we conclude that the two methods occupy regions near the perception and distortion optimization extremes of the trade-off boundary.  At high masking rates, the LM-RBM exhibits the perception-distortion trade-off in relation to our untrained methods, though not to the extent as MisGAN.  We view this as an indication that, as a generative model, the LM-RBM still optimizes inferred perceptual quality to some degree, and so must be subject to some increase in reconstruction distortion. 

MisGAN and the LM-RBM appear to consistently outline the locations of the perception and distortion extremes of the trade-off boundary.  The exact location of this boundary defines the limit of achievable performance of reconstruction methods.  Future work would be needed to determine the location of the trade-off boundary in missing data contexts.  In the complete data setting, Blau and Michaeli \cite{blau2018perception} demonstrate that the boundary may be traversed by training GANs with varying weights to reconstruction loss.   In the missing data setting, the true reconstruction loss remains inaccessible, so it is not clear whether such a traversal method can be adapted to settings where training data is also degraded.  A possible adaptation is to use the leave-one-out style reconstruction loss on a validation set or to factor in a prediction loss for side information (such as data labels).

The successful application of RBMs to reconstruction tasks also suggests the need for future work regarding reconstruction performance measures.  For our RBM-based reconstructions, the properties expressed by reconstruction-focused measures, like RMSE and FID, are determined by characteristics of the hidden layer activations.  Yet, despite being an integral part to the RBMs' reconstructions, the hidden layer representations are not themselves reconstructions of some ground truth. Therefore, an interesting future work would be to determine how one may measure the concept of perceptual quality based solely on intermediate inferences, such as those provided by the hidden layers of an RBM.

%


\balance
\bibliographystyle{IEEEbib}
\bibliography{refs}
\end{document}